\pdfoutput=1
\documentclass[11pt]{article}

\usepackage[]{acl}
\usepackage{enumitem}
\usepackage{algorithm}
\usepackage{algorithmicx}
\usepackage{amsmath}
\usepackage[noend]{algpseudocode}
\usepackage{booktabs}

\newcommand{\benchmark}{Advbench}
\newcommand{\advintent}{security}

\usepackage{color}
\usepackage{times}
\usepackage{latexsym}

\usepackage{graphicx}
\usepackage{multirow}
\usepackage[T1]{fontenc}
\usepackage{url}

\usepackage[utf8]{inputenc}

\usepackage{microtype}

\usepackage{comment}
\newif\ifshowcomment
    \showcommentfalse

\ifshowcomment
    \newcommand{\ganqu}[1]{\textcolor{purple}{[{ganqu: #1}]}}
    \newcommand{\yang}[1]{\textcolor{blue}{[yang: #1]}}
    
\else
    \newcommand{\todo}[1]{}
    \newcommand{\ganqu}[1]{}
    \newcommand{\yang}[1]{}
    \newcommand{\focus}[1]{}
\fi

\title{Why Should Adversarial Perturbations be Imperceptible? \\ Rethink the Research Paradigm in Adversarial NLP \\
\normalsize{\textcolor{red}{WARNING: This paper contains real-world cases which are offensive in nature.}}}

\author{
Yangyi Chen$^{1,2}$\thanks{\ \ Indicates equal contribution. Work done during internship at Tsinghua University.}\hspace{0.3em},
Hongcheng Gao$^{1,3*}$,
Ganqu Cui$^{1}$,
Fanchao Qi$^{1}$ \\
{\bf Longtao Huang$^{4}$,
Zhiyuan Liu$^{1,5}$\thanks{\ \ Corresponding Author.},
Maosong Sun$^{1,5}$\footnotemark[2]
}
\\ %\hspace{0.5em}
$^{1}$NLP Group, DCST, IAI, BNRIST, Tsinghua University, Beijing \\
$^{2}$University of Illinois Urbana-Champaign 
$^{3}$Chongqing University \\
$^{4}$Alibaba Group 
$^{5}$ IICTUS, Shanghai\\
{\tt yangyic3@illinois.edu,}
{\tt gaohongcheng2000@gmail.com}
}

\begin{document}

\maketitle

\begin{abstract}
Textual adversarial samples play important roles in multiple subfields of NLP research, including security, evaluation, explainability, and data augmentation.
However, most work mixes all these roles, obscuring the problem definitions and research goals of the security role that aims to reveal the practical concerns of NLP models.
In this paper, we rethink the research paradigm of textual adversarial samples in \advintent~scenarios.
We discuss the deficiencies in previous work and propose our suggestions that the research on the \textbf{S}ecurity-\textbf{o}riented \textbf{ad}versarial \textbf{NLP} \textbf{(SoadNLP)} should:
(1) evaluate their methods on \advintent~tasks to demonstrate the real-world concerns;
(2) consider real-world attackers' goals, instead of developing impractical methods. 
To this end, we first collect, process, and release a \advintent~datasets collection \textbf{\benchmark}.
Then, we reformalize the task and adjust the emphasis on different goals in SoadNLP. 
Next, we propose a simple method based on heuristic rules that can easily fulfill the actual adversarial goals to simulate real-world attack methods.
We conduct experiments on both the attack and the defense sides on \benchmark. 
Experimental results show that our method has higher practical value, indicating that the research paradigm in SoadNLP may start from our new benchmark.
All the code and data of Advbench can be obtained at \url{https://github.com/thunlp/Advbench}.

\end{abstract}

\section{Introduction}

Natural language processing (NLP) models based on deep learning have been employed in many real-world applications \cite{badjatiya2017deep,zhang2018deep,niklaus2018survey,han2021pre}. 
Meanwhile, there is a concurrent line of research on textual adversarial samples that are intentionally crafted to mislead models' predictions \citep{samanta2017towards, papernot2016crafting}. 
Previous work shows that textual adversarial samples play important roles in multiple subfields of NLP research.
We categorize and summarize the roles in Table~\ref{role}.

We argue that the problem definitions, including priorities of goals and experimental settings, are different, considering the different roles of adversarial samples.
However, most previous work in adversarial NLP mixes all different roles,  including the security role of revealing real-world concerns of NLP models deployed in \advintent~scenarios. 
This leads to inconsistent problem definitions and research goals with real-world cases.
As a consequence, although most existing work on textual adversarial attacks claims that their methods reveal the security issues, they often follow a security-irrelevant research paradigm.
To fix this problem, we focus on the security role and try to refine the research paradigm for future work in this direction.

\begin{table}[]
\centering
\renewcommand\arraystretch{0.98}
\resizebox{.5\textwidth}{!}{\begin{tabular}{ll}
\toprule[1.2pt]
\textbf{Role}  & \textbf{Explanation}                                                                                                                                           \\ \midrule
Security       &\begin{tabular}[c]{@{}l@{}}Adversarial samples can reveal the practical concerns of NLP\\models deployed in security situations.\end{tabular}      \\ \midrule

Evaluation     & \begin{tabular}[c]{@{}l@{}}Adversarial samples can be employed to benchmark models’\\robustness to out-of-distribution data (diverse user inputs).\end{tabular} \\ \midrule

Explainability & \begin{tabular}[c]{@{}l@{}}Adversarial samples can explain part of the models' decision\\processes.\end{tabular}                                                 \\ \midrule
Augmentation   & \begin{tabular}[c]{@{}l@{}}Adversarial training based on adversarial samples augmenta-\\tion can improve performance and robustness.  \end{tabular}  \\ \bottomrule[1.2pt]

\end{tabular}
}

\caption{\label{role} Roles of textual adversarial samples.}

\end{table}

\begin{table*}[]
\centering
\resizebox{.95\textwidth}{!}{
\begin{tabular}{l|l}
\toprule[1.2pt]
Original                               & {\color[HTML]{000000} I was all over the fucking place because the toaster had tits.}              \\ \midrule
PWWS     \cite{ren-etal-2019-generating}                              & {\color[HTML]{000000} I was all over the \color[HTML]{FE0000}bally topographic \color[HTML]{000000}because the \color[HTML]{FE0000}wassailer have breast.}     \\ \midrule
\multicolumn{1}{l|}{Real-World Attack} & {\color[HTML]{000000} I was all over the \color[HTML]{FE0000}fuc king \color[HTML]{000000} place because the toaster had tits. \color[HTML]{FE0000}!!!peace peace peace} \\ \bottomrule[1.2pt]
\end{tabular}
}
\caption{\label{case_real_attack} 
Comparison between the real-world attack and the method proposed in the NLP community. 
Obviously, the real-world attack method is easier to implement and preserves the adversarial meaning better. }

\end{table*}
%We identify some deficiencies in previous works. 
%Most existing works on textual adversarial attacks claim that their attack methods reveal the security issues. 
%However, we find two common shortcomings of previous works.
\looseness=-1
There are two core issues about why previous textual adversarial attack work can hardly help real-world security problems.
First, most work don't consider \advintent~tasks and datasets \cite{ren-etal-2019-generating, zang-etal-2020-word} (See Table~\ref{previousworks}). 
Some irrelevant tasks like sentiment analysis and natural language inference are often involved in the evaluation instead. 
Second, they don't consider real-world attackers' goals and make unrealistic assumptions or add unnecessary restrictions (e.g., imperceptible requirement) to the adversarial perturbations \cite{li-etal-2020-bert-attack, garg-ramakrishnan-2020-bae}. 
Consider the case where attackers want to bypass the detection systems to send an offensive message to the web. 
They can only access the decisions (e.g., pass or reject) of the black-box detection systems without the concrete confidence scores. 
And their adversarial goals are to convey the offensive meaning and bypass the detection systems. 
So, there is no need for them to make the adversarial perturbations imperceptible, as supposed in previous work. 
See Table~\ref{case_real_attack} for an example. 
Besides, most methods have the inefficiency problem (i.e. high query times and long-running time), which makes them less practical and may not be a good choice for attackers in the real world.
We refer readers to Section~\ref{discuss:reflect} for a further discussion about previous work.

% \yang{Whether to add the paragraph here?} One thing to note is that we don't discredit existing works and we appreciate their contributions in this field on remaining values of adversarial samples. For example, consider the PWWS attack algorithms \cite{}. It can find adversarial samples based on synonym substitutions and greedy search algorithm. If crafted correctly, this algorithms can fulfill all remaining research values of textual adversarial samples. 

To address the issue of security-irrelevant evaluation benchmark, we first summarize five \advintent~tasks and search corresponding open-source datasets. 
We collect, process, and release these datasets as a collection named \textbf{\benchmark}~to facilitate future research. 
%To address the issue of ill-defined problem definition, we start from the reality and rethink real-world attackers' goals. 
%We reformalize the task of textual adversarial attack considering the security role of adversarial samples. 
%We mainly focus on adjusting the emphasis on different adversarial goals, referring to the intention of real-world attackers.
To address the issue of ill-defined problem definition, we refer to the intention of real-world attackers to reformalize the task of textual adversarial attack and adjust the emphasis on different adversarial goals. 
Further, to simulate real-world attacks, we propose a simple attack method based on heuristic rules that are summarized from various sources, which can easily fulfill the actual attackers' goals.

We conduct comprehensive experiments on \benchmark~to evaluate methods proposed in the NLP community and our simple method.  
Experimental results overall demonstrate the superiority of our method, considering the attack performance, the attack efficiency, and the preservation of adversarial meaning (validity). 
We also consider the defense side and show that the SOTA defense method cannot handle our simple heuristic attack algorithm.
The overall experiments indicate that the research paradigm in SoadNLP may start from our new benchmark.

To summarize, the main contributions of this paper are as follows:
\begin{itemize} [topsep=1pt, partopsep=1pt, leftmargin=12pt, itemsep=-2pt]
	\item We collect, process, and release a \advintent~datasets collection \benchmark.
	\item We reconsider the attackers' goals and reformalize the task of textual adversarial attack in security scenarios.  
	\item We propose a simple attack method that fulfills the actual attackers' goals to simulate real-world attacks, which can facilitate future research on both the attack and the defense sides.
% 	\item On the whole, we identify the deficiencies in previous works in adversarial NLP and propose our refinements, pointing out the potential research directions in this filed (See Section~\ref{discuss:reflect}). 
\end{itemize}

\section{\benchmark~Construction}

\subsection{Motivation}
\looseness=-1
We first survey previous works of adversarial attacks in NLP about the tasks and datasets they consider in their experiments (See Table~\ref{previousworks}).
We find that most tasks consider in their work are not \advintent-relevant (e.g., sentiment analysis). 
So, the real-world concerns revealed in their experiments are not well reflected in reality when there is a lack of \advintent~evaluation benchmark.
To this end, we suggest future researchers evaluate their methods on \advintent~tasks to demonstrate real-world harmfulness and practical concerns. 
Thus, a \advintent~datasets collection is needed to facilitate future research.

\subsection{Tasks}
\looseness=-1
We summarize 5 \advintent~tasks, including misinformation, disinformation, toxic, spam, and sensitive information detection. 
The task descriptions and our motivation to choose these tasks are given in Appendix~\ref{appendix:task}.
Due to the label-unbalanced issue of some datasets, 
we will release both our processed balanced and unbalanced datasets.
The datasets statistics are listed in Table~\ref{dataset_description}. 
All datasets are processed through the general pipeline including the removal of duplicate, missing, and unusual values.
% Note that for all \advintent~tasks, only two categories are needed, namely the unfavorable and normal cases. 

% so for tasks with more than two categories, we will combine them into two categories

\subsubsection{Misinformation}
%  misinformation 
\paragraph{LUN.} Our LUN dataset is built on
the Labeled Unreliable News Dataset \cite{rashkin2017truth} consisting of articles from news media and human annotations of fact-checking. 
We merge the satirical news from the Onion, hoax from the American News, and propaganda from the Activist Report into one category labeled as untrusted.
And the articles collected from Gigaword News are labeled as trusted. 
Considering there is too little data in the original testing set, we mix the original training and testing set and re-partition by 7:3.

\paragraph{SATNews.} The Satirical News Dataset~\cite{yang2017satirical} is a collection of satirical and verified news. 
% Satirical news is also a form of fact concealment. 
The satirical news articles are collected from 14 websites that explicitly declare that they are offering satire.
The verified news articles are collected from major news outlets\footnote{CNN, DailyMail, WashingtonPost, NYTimes, The Guardian, and Fox.} and Google News using FLORIN~\cite{liu2015florin}. 
The original training set and validation set are merged as our training set and the testing set remains unchanged.

% \paragraph{EMNLP2017 Satirical News}
% \gao{name}
% \paragraph{CNFN} CompareNet FakeNews   Trusted Satire Hoax Propaganda
% \gao{name}
\subsubsection{Disinformation}

% \paragraph{Yelp}
% % Amazon Review Data 
% \paragraph{ARD2018} The Amazon Review Data 2018 \cite{ni2019justifying} contains 233.1 million product reviews with corresponding product metadata from Amazon in the range May 1996 - Oct 2018, which is an updated version of the Amazon review dataset released in 2014 \cite{he2016ups,mcauley2015image}. The ARD2018 has 12 categories and two data forms: complete review data and the "small" subsets. The "small" reviews subsets reduced to extract the 5-core has been collected and divided into training set and validation set by the ratio of 7:3.  Besides, only review text and verification information(verified:true/false) has been retained in our dataset.

\paragraph{Amazon-LB.} 
\looseness=-1
The Amazon Luxury Beauty Review dataset is a review collection of the Luxury Beauty category in Amazon with verification information in Amazon Review Data (2018)~\cite{ni2019justifying}. 
The Amazon Review Data (2018) is an updated version of the Amazon Review Dataset~\cite{he2016ups,mcauley2015image} released in 2014, which contains 29 types of data for different scenarios.  
We extract the Luxury Beauty data from "small" subsets that are reduced from full sets due to the appropriate quantity and diversity of this category. 
% to extract the K-core so that each of the remaining users and items have k reviews each
We only keep content and label (whether the content is verified or not) of the review and split the data into training and testing set with a ratio of 7:3.

% The Amazon Luxury Beauty Review dataset is review collection of the category Luxury Beauty in Amazon with verification information, which is a subset of Amazon Reviews 2018 \cite{ni2019justifying}  Amazon Review 2018 is an updated version of the Amazon review dataset released in 2014 \cite{he2016ups,mcauley2015image} and "Small" subsets have been reduced to extract the k-core. 

\paragraph{CGFake.} The Computer-generated Fake Review Dataset \cite{SALMINEN2022102771} contains label-balanced product reviews with two categories: original reviews (presumably human-created and authentic) and computer-generated fake reviews. 
The computer-generated fake review is a new type of disinformation that employs computer technology to generate fake samples to mislead humans. 
This dataset is split into training and testing set the same as the original paper.

\subsubsection{Toxic}

% Toxic detection is a word-sensitive task which means the alteration of key words can flip the expected label. 
% So that sentences in such task can easily fool the detection system through word replacement, such as replacing fuck with fack.

\paragraph{HSOL.}
The Hate Speech and Offensive Language
Dataset~\cite{hateoffensive} contains more than 200k labeled tweets which are searched by Twitter API. 
The original dataset is classified into three categories: hate speech, offensive but not hate speech, or normal. 
We combine hate and offensive speech into one category labeled "hate" and the others are labeled as "non-hate".

% \paragraph{Twitter} The Twitter \cite{founta2018large}  abusive behavior dataset contains 80k tweets annotated concerning Inappropriate Speech (more particularly in matters of Abusive and Hateful speech) as well as Normal and Spam. The original dataset has 8 classes and we collected the dataset from RIPPLe \cite{kurita20acl} which has been processed into 2 classes: toxicity and non-toxic speech with the ratio of 1:1.68.
% \paragraph{OLID}  The Offensive Language Identification
% Dataset \cite{zampieri-etal-2019-predicting} is a text dataset with over 14,000 English twee
% tweets annotated for offensive content using
% a fine-grained three-layer annotation scheme. It featured three sub-tasks. In sub-task A, the goal was to discriminate between offensive and non-offensive posts. In sub-task B, the focus was on the type of offensive content in the post. Finally, in sub-task C, systems had to detect the target of the offensive posts. We collected the proccessed data released by RIPPLe \cite{kurita20acl} which has two labels(non-offensive/offensive) with the ratio of almost 2:1.
\paragraph{Jigsaw2018.} 
The Jigsaw2018\footnote{This dataset is available in \href{https://www.kaggle.com/c/jigsaw-toxic-comment-classification-challenge}{Kaggle}.} is a competition dataset of Toxic Comment Classification Challenge in Kaggle. 
This dataset includes plentiful Wikipedia comments. 
And the comments are labeled by human annotators for toxic behavior with two categories: toxic and non-toxic. 

% The validation samples will be used as thetest set in our work.

\subsubsection{Spam}

\paragraph{Enron.} 
The Enron\footnote{\url{http://www2.aueb.gr/users/ion/data/enron-spam/}}~\cite{metsis2006spam} is a corpus of emails split into two categories: legitimate and spam.  
There are six subsets in the dataset.
Each subset contains non-spam messages from a user in the Enron corpus.
And each non-spam message is paired with one of the three spam collections including the SpamAssassin corpus and the Honeypot project\footnote{\url{https://www.projecthoneypot.org/}}, Bruce Guenter's spam collection\footnote{\url{http://untroubled.org/spam/}}, and the spam collected by \citet{metsis2006spam}.
We mix all the datasets and split them into training and testing sets. 
We only keep the content of each email without other information such as subject and address.

\paragraph{SpamAssassin.} 
The SpamAssassin\footnote{\url{https://spamassassin.apache.org/old/publiccorpus/}} is a collection of emails consisting of three categories: easy-ham, hard-ham, and spam. 
We merge easy-ham and hard-ham as the ham class.
Then we mix all samples and split them equally into training and testing sets because of the lack of data. 
For each email, we preprocess it the same as Enron.

\subsubsection{Sensitive Information}

% Complex sensitive information is characterized by the fact that words are sensitive or not sensitive depending on their context. 
% The detection of sensitive information is very easy to be misjudged, making this task not only important but also very meaningful.

\paragraph{EDENCE.} EDENCE~\cite{DVN/WRL7ZS_2019} contains samples with auto-generated parsing-tree structures in the Enron corpus. 
The annotated labels come from the TREC LEGAL~\cite{tomlinson2010learning,cormack2010overview} labels for Enron documents.
We restore the tree-structured samples to normal texts and map sensitive information labels back to each sample. 
Then we combine the training and validation sets as our training set, and the testing set remains unchanged.

\paragraph{FAS.} FAS~\cite{DVN/C9PYZM_2019} also contains samples with parsing-tree structures built from Enron corpus and is modified for sensitive information detection by using TREC LEGAL labels annotated by domain experts. 
The samples in FAS are compliant with Financial Accounting Standards 3 and are preprocessed in the same way as EDENCE in our work.
% FAS has sentences where Enron claims compliance with Financial Accounting Standards 3.

\section{Task Formalization}

\subsection{Motivation}
\looseness=-1
In our survey, we find that the current problem definition and research goals considering the security role of adversarial samples to reveal practical concerns are ill-defined and ambiguous. 
We attribute this to the failure of distinguishing several roles of adversarial samples (See Table~\ref{role}). 
The problem definitions are different considering the different roles of adversarial samples. 
For example, when adversarial samples are adopted to augment existing datasets for adversarial training, we may aim for high-quality samples. Thus, the minor perturbations restriction is important.
On the contrary, when it comes to the security side, we should focus more on the preservation of adversarial meaning and attack efficiency instead of the imperceptible perturbations.
% while samples quality and minor perturbations are important when adversarial samples are employed to augment existing datasets for adversarial training,  when it comes to the security side instead of the imperceptible perturbations. 
See section~\ref{discuss:reflect} for a further discussion. 

Thus, we need to separate the research on different roles of adversarial samples. 
On the security side, most work doesn't consider realistic situations and the actual adversarial goals, which may result in unrealistic assumptions or unnecessary restrictions when developing attack or defense methods.
To make the research in this field more standardized and in-depth, reformalization of this problem needs to be conducted. \textbf{Note that we focus on the security role of textual adversarial samples in this paper.}

\subsection{Formalization}
\paragraph{Overview.}
Without loss of generality, we consider the text classification task. 
Given a classifier $f:\mathcal{X} \to \mathcal{Y}$ that can make correct prediction on the original input text \textbf{x}:
\begin{equation}
    \mathop{\arg\max}\limits_{y_i \in \mathcal{Y}} \mathcal{P}(y_i|\textbf{x}) = y_{true},
\end{equation}
where $y_{true}$ is the golden label of $\textbf{x}$. 
The attackers will make perturbations $\delta$ to craft an adversarial sample  $\textbf{x}^*$ that can fool the classifier:
\begin{equation}
    \mathop{\arg\max}\limits_{y_i \in \mathcal{Y}} \mathcal{P}(y_i|\textbf{x}^*) \neq y_{true}, 
   \quad \textbf{x}^* = \textbf{x} + \delta
\end{equation}

\paragraph{Refinement.}
The core part of adversarial NLP is to find the appropriate perturbations $\delta$. 
We identify four deficiencies in the common research paradigm on SoadNLP.

(1) Most attack methods iteratively search for better $\delta$ relying on the accessibility to the victim models' confidence scores or gradients \cite{alzantot-etal-2018-generating, ren-etal-2019-generating, zang-etal-2020-word, li-etal-2020-bert-attack}. 
However, this assumption is unrealistic in real-world \advintent~tasks (e.g., hate-speech detection). 
We argue that the research in adversarial NLP considering the practical concerns should focus on the decision-based setting, where only the decisions of the victim models can be accessed.

\looseness=-1
(2) Previous work attempts to make $\delta$ imperceptible by imposing some restrictions on the searching process, like ensuring that the cosine similarity of adversarial and original sentence embeddings is higher than a threshold \cite{li-etal-2020-bert-attack, garg-ramakrishnan-2020-bae}, or considering the adversarial samples' perplexity \cite{qi-etal-2021-mind}.  
However, why should adversarial perturbations be imperceptible? 
The goals of attackers are to (1) bypass the detection systems and (2) convey the malicious meaning. 
So, the attackers only need to preserve the adversarial contents (e.g., the hate speech in messages) no matter how many perturbations are added to the original sentence to bypass the detection systems (Consider Table~\ref{case_real_attack}). 
Thus, we argue that these constraints are unnecessary and the quality of adversarial samples is a secondary consideration.

(3) Adversarial attack based on word substitution or sentence paraphrase is the most widely studied. 
However, current attack algorithms are very inefficient and need to query victim models hundreds of times to craft adversarial samples, which makes them unlikely to happen in reality\footnote{Some work tries to address this issue but the effect is limited \cite{zang2020learning, chen-etal-2021-multi}.}.
We argue that adversarial attacks should be computation efficient, both in the running time and the query times to the victim models, to better simulate the practical situations.

(4) There is a bunch of work assuming that the attackers are experienced NLP practitioners and incorporate external knowledge base \cite{ren-etal-2019-generating, zang-etal-2020-word} or NLP models \cite{li-etal-2020-bert-attack, qi-etal-2021-mind} into their attack algorithms. 
However, everyone can be an attacker in reality. 
Consider the hate-speechers in social platforms. 
They often try different heuristic strategies to escape detection without any knowledge in NLP (See Appendix~\ref{appendix:real_word_case} for cases). 
Besides, the research in the security community confirms that real-world attackers only use some simple heuristic attack methods to propagate illicit online promotion instead of the complicated ones proposed in the computer vision domain \cite{yuan2019stealthy}. 
We argue that besides the professional approaches that have been extensively studied, the research on adversarial attack and defense should also pay some attention to simple and heuristic methods that many real-world attackers are currently employing.

\begin{table}[]
\centering
\resizebox{.49\textwidth}{!}{
\begin{tabular}{lll}
\toprule[1.2pt]
\textbf{Goal}      & \textbf{Metric}                             & \textbf{Priority}           \\ \midrule
Fool Detector      & {\color[HTML]{000000} Attack Success Rate}  & {\color[HTML]{000000} First} \\
Preserve Adversarial Meaning           & {\color[HTML]{000000} Validity}     & {\color[HTML]{000000} First} \\
Reduce Computation and Query & {\color[HTML]{000000} Query Time}          & {\color[HTML]{000000} First} \\
Minor Perturbations  & {\color[HTML]{000000} Levenstein Distance}        & {\color[HTML]{000000} Second}  \\
Adversarial Sample Quality           & {\color[HTML]{000000} PPL \& Grammar Error} & Second                         \\ \bottomrule[1.2pt]
\end{tabular}
}
\caption{\label{metrics} The priority of adversarial goals and corresponding evaluation metrics. \yang{Future work may include discussion how we evaluate attack methods in security tasks, feature-level perturbations}}

\end{table}
In general, we make two suggestions for future research, including considering the decision-based experimental setting and the attack methods that are free of expertise. 
Besides, we adjust the emphasis on different adversarial goals, corresponding to the real-world attack situations (See Table~\ref{metrics}).
Note that the validity requirement (preservation of adversarial meaning) of adversarial samples is task-specific and we discuss it in Appendix~\ref{appendix:validity_def}. Compared to previous work, we set different priorities for different goals and put more emphasis on the preservation of adversarial meaning and the computation efficiency, while down-weighting the attention to minor perturbations and sample quality.

\textbf{Note that we don't convey the meaning that the quality of adversarial samples is not important.} 
For example, spam emails and fake news will obtain more attacker-expected feedback if they are more fluent and look more natural. 
Our intention in this paper is to decrease the priority of the secondary adversarial goals when there exists a trade-off among all adversarial goals, to better simulate real-world attack situations. 

\subsection{Our Method}
To simulate the adversarial strategies employed by real-world attackers, we also propose a simple method named \textbf{ROCKET} (\textbf{R}eal-w\textbf{O}rld atta\textbf{CK} based on h\textbf{E}uris\textbf{T}ic rules) that can fulfill the actual adversarial goals.
Our algorithm can be divided into two parts, including heuristic perturbation rules and the black-box searching algorithm.

\begin{table}[]
\centering
\resizebox{.5\textwidth}{!}{
% \begin{tabular}{l|l|l}
% \hline
% \multicolumn{1}{c|}{Rule} & \multicolumn{1}{c|}{{\color[HTML]{000000} Description}} & \multicolumn{1}{c}{Example}                     \\ \hline
% (1) Insert Space           & {\color[HTML]{000000} Randomly insert a space}          & foolish -\textgreater foo lish                  \\ \hline
% (2) Insert Irrelevant      & {\color[HTML]{000000} Randomly insert a character}      & foolish -\textgreater foo\textasciicircum{}lish \\ \hline
% (3) Delete                 & Randomly delete a character                             & foolish -\textgreater foolih                    \\ \hline
% (4) Swap                   & Randomly swap two adjacent characters                   & foolish -\textgreater fooilsh                   \\ \hline
% (5) Substitute             & Randomly substitute a character                          & foolish -\textgreater foo1ish                   \\ \hline
% (6) Add Distractor         & Add distracting words.                    & fuck -\textgreater fuck peace!!             \\ \hline
% \end{tabular}
\begin{tabular}{lll}
\toprule[1.2pt]
\textbf{Rule}         & {\color[HTML]{000000} \textbf{Description}}        & \textbf{Example}                                \\ \hline
(1) Insert Space      & {\color[HTML]{000000} Randomly insert a space}     & foolish -\textgreater ~foo lish                  \\
(2) Insert Irrelevant & {\color[HTML]{000000} Randomly insert a character} & foolish -\textgreater       ~foo\textasciicircum{}lish \\
(3) Delete            & Randomly delete a character                        & foolish -\textgreater ~foolih                    \\
(4) Swap              & Randomly swap two adjacent characters              & foolish -\textgreater ~fooilsh                   \\
(5) Substitute        & Randomly substitute a characer                     & foolish -\textgreater ~foo1ish                   \\
(6) Add Distractor    & Add distracting sentence at the end                & fuck! -\textgreater ~fuck peace!!                 \\ \bottomrule[1.2pt]
\end{tabular}
}
\caption{\label{perturbation_rules} Heuristic perturbation rules.}

\end{table}
\paragraph{Perturbation Rules.} 
\looseness=-1
To make our heuristic perturbation rules better simulate real-world attackers, we survey and summarize common perturbations rules from several sources, including 
(1) real adversarial user data (some cases are shown in Appendix~\ref{appendix:real_word_case}), 
(2) senior practitioners' experience, 
(3) papers in the NLP community~\citep{jia-liang-2017-adversarial, ebrahimi2017hotflip},
(4) reports of adversarial competitions, and (5) our intuition from the attackers' point of view. 
We filter the rules and retain only those that are \textbf{common}, \textbf{computation efficient}, and \textbf{easy to implement without any external knowledge} (See Table~\ref{perturbation_rules}).
The big difference between ROCKET and previous methods (e.g., DeepWordBug) is its easy-to-implement property, which allows it to be actually employed by real-world attackers without any external knowledge.

We now specify how we find distracting words (rule-6).
For each task, we first gather some realistic data and obtain the words that occur relatively more in attacker-specified labeled samples (e.g., non-spam in the spam detection task) by calculating word frequency. 
Then we heuristically select distracting words that will not interfere with the original task. 
Finally, we add an appropriate amount of selected words at the beginning or end of the original sentence, ensuring that the semantics of the sentence will not be affected.

\vspace{-1pt}
\paragraph{Searching Algorithm.} 
We need to heuristically apply perturbations rules to search adversarial samples in the black-box setting because only victim models' decisions are available.  
We first apply rule-6 to the original sentence and filter stop words to get the semantic word list $L$ of the modified sentence. 
Then we repeat the word perturbation process while not fooling the victim model. 
Specifically, one iteration of the word perturbation process starts by first sampling a batch of words $w$ from $L$. Repeat the process of sampling actions $r$ from rule-1 to rule-5 for each word in $w$ and query the victim model until the threshold is reached or the attack succeeds. 
Then $w$ is removed from $L$.
% The word batch size and the threshold are hyper-parameters and set according to the characteristic of the specific task.
% The whole process is shown as pseudocode in Appendix. 

\begin{table*}[]
\centering
\resizebox{0.95\textwidth}{!}{
\begin{tabular}{lcccccccccccccc}
\toprule[1.2pt]
\textbf{Task}                                                 & \multicolumn{2}{c}{\textbf{Misinformation}} &  & \multicolumn{2}{c}{\textbf{Disinformation}} &  & \multicolumn{2}{c}{\textbf{Toxic}} &  & \multicolumn{2}{c}{\textbf{Spam}} &  & \multicolumn{2}{c}{\textbf{Sensitive Information}} \\ \hline
\multicolumn{1}{l|}{\multirow{2}{*}{\textbf{Method~|~Dataset}}} & \multicolumn{2}{c}{\underline{LUN}}           &  & \multicolumn{2}{c}{\underline{Amazon-LB} }              &  & \multicolumn{2}{c}{\underline{HSOL} }          &  & \multicolumn{2}{c}{\underline{SpamAssassin}}  &  & \multicolumn{2}{c}{\underline{EDENCE}}             \\
\multicolumn{1}{l|}{}                                         & ASR(\%)        & Query            &  & ASR(\%)             & Query                 &  & ASR(\%)         & Query            &  & ASR(\%)        & Query            &  & ASR(\%)           & Query              \\ \hline
\multicolumn{1}{l|}{TextFooler}                               & 0.4            & 1294.38          &  & 9.0                 & 740.42                &  & 10.4            & 78.46            &  & 0.2            & 961.88           &  & 23.9              & 94.67              \\
\multicolumn{1}{l|}{PWWS}                                     & 1.3            & 1707.19          &  & 18.8                & 1019.91               &  & 9.9             & 107.23           &  & 0.3            & 1308.50          &  & 46.0              & 129.68             \\
\multicolumn{1}{l|}{BERT-Attack}                              & 7.0            & 3966.60          &  & \textbf{43.0}       & 1625.37               &  & 56.8            & 139.14           &  & \textbf{2.2}   & 4336.18          &  & \textbf{90.3}     & 140.98             \\
\multicolumn{1}{l|}{SememePSO(maxiter=100)}                   & 0.9            & 2020.85          &  & 23.8                & 1627.97               &  & 66.9            & 233.11           &  & 0.9            & 1945.74          &  & 79.6              & 231.17             \\
\multicolumn{1}{l|}{DeepWordBug(power=5)}                     & 0.1            & \textbf{287.04}  &  & 9.3                 & \textbf{162.37}       &  & 56.4            & 21.43            &  & 0.1            & 263.84           &  & 22.9              & 26.06              \\
\multicolumn{1}{l|}{DeepWordBug(power=25)}                    & 0.2            & 287.04           &  & 12.4                & 162.41                &  & \textbf{85.4}   & 21.72            &  & 0.0              & 263.84           &  & 79.9              & 26.63              \\
\multicolumn{1}{l|}{\textbf{ROCKET}}                          & \textbf{7.2}   & 300.38           &  & 38.7                & 218.69                &  & 72.4            & \textbf{18.50}   &  & 1.1            & \textbf{60.09}   &  & 84.5              & \textbf{20.93}     \\ \hline
\textbf{Acc.(\%)}                                             & \multicolumn{2}{c}{\textbf{99.2}} &  & \multicolumn{2}{c}{\textbf{92.1}}           &  & \multicolumn{2}{c}{\textbf{95.7}}  &  & \multicolumn{2}{c}{\textbf{99.3}} &  & \multicolumn{2}{c}{\textbf{96.3}}      \\ \hline
\multicolumn{1}{c}{}                                          &                &                  &  &                     &                       &  &                 &                  &  &                &                  &  &                   &                    \\ \hline
\multicolumn{1}{l|}{\multirow{2}{*}{\textbf{Method~|~Dataset}}} & \multicolumn{2}{c}{\underline{SATNews}}       &  & \multicolumn{2}{c}{\underline{CGFake}}                  &  & \multicolumn{2}{c}{\underline{Jigsaw2018}}     &  & \multicolumn{2}{c}{\underline{Enron}}         &  & \multicolumn{2}{c}{\underline{FAS} }               \\
\multicolumn{1}{l|}{}                                         & ASR(\%)        & Query            &  & ASR(\%)             & Query                 &  & ASR(\%)         & Query            &  & ASR(\%)        & Query            &  & ASR(\%)           & Query              \\ \hline
\multicolumn{1}{l|}{TextFooler}                               & 2.9            & 1889.32          &  & 18.2                & 360.13                &  & 12.5            & 201.72           &  & 0.1            & 682.40           &  & 17.4              & 130.73             \\
\multicolumn{1}{l|}{PWWS}                                     & 1.2            & 2565.37          &  & 69.0                & 489.78                &  & 20.2            & 268.87           &  & 0.0              & 928.58           &  & 36.5              & 177.18             \\
\multicolumn{1}{l|}{BERT-Attack}                              & \textbf{30.6}  & 5102.34          &  & 94.6                & 400.61                &  & 40.4            & 450.67           &  & 1.4            & 2954.86          &  & \textbf{92.4}     & 305.59             \\
\multicolumn{1}{l|}{SememePSO(maxiter=100)}                   & 4.2            & 2217.34          &  & 67.2                & 689.46                &  & 51.9            & 539.25           &  & 1.0            & 1724.70          &  & 61.4              & 506.82             \\
\multicolumn{1}{l|}{DeepWordBug(power=5)}                     & 2.5            & 430.59           &  & 41.7                & 75.00                 &  & 35.9            & \textbf{45.71}   &  & 0.0              & 182.27           &  & 40.8              & \textbf{39.91}     \\
\multicolumn{1}{l|}{DeepWordBug(power=25)}                    & 1.9            & 430.59           &  & 68.8                & 75.28                 &  & 57.6            & 45.92            &  & 0.0              & 182.27           &  & 77.6              & 40.27              \\
\multicolumn{1}{l|}{\textbf{ROCKET}}                          & 4.4            & \textbf{324.30}  &  & \textbf{97.2}       & \textbf{37.11}        &  & \textbf{64.2}   & 78.85            &  & \textbf{6.5}   & \textbf{56.92}   &  & 82.0              & 52.77              \\ \hline
\textbf{Acc.(\%)}                                             & \multicolumn{2}{c}{\textbf{96.6}} &  & \multicolumn{2}{c}{\textbf{99.1}}           &  & \multicolumn{2}{c}{\textbf{95.5}}  &  & \multicolumn{2}{c}{\textbf{99.7}} &  & \multicolumn{2}{c}{\textbf{97.8}}      \\ \bottomrule[1.2pt]
\end{tabular}}

\caption{\label{main_results} Results of first priority metrics considering the attack performance and the attack efficiency. }

\end{table*}
\section{Experiments}

\subsection{Experimental Settings}
\paragraph{Dataset and Victim Model.}
We choose BERT-base \cite{devlin-etal-2019-bert} as the victim model and evaluate attack methods on our \benchmark.

\paragraph{Evaluation Metrics.}
We evaluate the attack methods considering
first priority goals, including attack success rate, attack efficiency, and validity, and second priority goals, including perturbation degree, and quality. 
(1) Attack success rate (ASR) is defined as the percentage of successful adversarial samples.
(2) Validity is measured by human annotators.
The annotation details are in Appendix~\ref{appendix:human_eval_details}. 
(3) Attack efficiency (Query) is defined as the average query times to the victim models when crafting adversarial samples. 
(4) Perturbation degree is measured by Levenstein distance.
(5) Quality is measured by the relative increase of perplexity and absolute increase of grammar errors when crafting adversarial samples.

\subsection{Baseline Methods}
We implement existing attack methods proposed in the NLP community using the NLP attack package OpenAttack \cite{zeng-etal-2021-openattack}. 
We comprehensively compare our simple method with five representative and strong attack models including 
(1) TextFooler \cite{jin2020bert}, 
(2) PWWS \cite{ren-etal-2019-generating}, 
(3) BERT-Attack \cite{li-etal-2020-bert-attack}, 
(4) SememePSO \cite{zang-etal-2020-word},
and (5) DeepWordBug \cite{gao2018black}. 
Specifically, we implement these methods in the black-box setting, where only the decisions can be accessed.

\subsection{Experimental Results} 
The experimental details can be found in Appendix~\ref{appendix:experimental_details}.

\begin{table}[]
\centering
\resizebox{.45\textwidth}{!}{\begin{tabular}{lcc}
\toprule[1.2pt]
\textbf{Method~|~Task} & \multicolumn{1}{l}{\textbf{Disinformation}} & \multicolumn{1}{l}{\textbf{Toxic}} \\ \hline
TextFooler              & 1.71                                        & 0.87                               \\
PWWS                    & 1.27                                        & 0.94                               \\
BERT-Attack             & 0.45                                        & 0.35                               \\
SememePSO(maxiter=100)  & 1.56                                        & 0.69                               \\
DeepWordBug(power=5)    & \textbf{2.00}                               & 1.45                               \\
DeepWordBug(power=25)   & \textbf{2.00}                               & 1.03                               \\
\textbf{ROCKET}         & 1.78                                        & \textbf{1.98}                      \\ \bottomrule[1.2pt]
\end{tabular}
}

\caption{\label{human_validity} The validity scores. 
The upper bound is \textbf{2}, which means that all selected adversarial samples preserve adversarial meaning.}

\end{table}
% \yang{add analysis about the dataset-specific results, e.g. word or sentence sensitive}
\paragraph{First Priority Metrics. }
We list the results of attack success rate and average query times in Table~\ref{main_results}. 
Our findings are as follows:
\begin{itemize} [topsep=1pt, partopsep=1pt, leftmargin=12pt, itemsep=-2pt]
    \item Considering all previous attack methods, we find that it's extremely hard to craft adversarial samples in some tasks (e.g., Misinformation, Spam). 
    And the attack performances of all methods drop compared to the results in original papers\footnote{The results of attack performance are actually overestimated if considering the validity of adversarial samples.}.
    We attribute this to the tough decision-based attack setting and the distinct features in these \advintent~tasks (the victim model achieves high accuracy on all these datasets). 

    \item Most previous methods are inefficient when launching adversarial attacks. Usually, they need to query the victim model hundreds of times to craft a successful adversarial sample.
    
    \item Our simple ROCKET shows superiority overall considering the attack performance and attack efficiency on \benchmark.
    
\end{itemize}
To further demonstrate the efficiency of ROCKET, we restrict the maximum query times to the victim model and test the attack success rate on Amazon-LB, HSOL, and EDENCE. 
The results are shown in Figure~\ref{fig:querytime}. We conclude that ROCKET shows stronger attack performance when the query time is restricted, which is more consistent with real-world situations.

We also conduct a human evaluation on the validity of adversarial samples (See Table~\ref{human_validity}). The details of the human evaluation process are described in Appendix~\ref{appendix:human_eval_details}. 
We conclude that character-level perturbations (e.g., DeepWordBug) can preserve adversarial meaning to the greatest extent possible while strong word-level attacks (e.g., BERT-Attack) seriously destroy the original adversarial meaning, which we suspect is caused by very uncommon words substitution (See Table~\ref{case_real_attack}). 
Besides, ROCKET achieves overall great validity compared to baselines. 

Note that ROCKET is designed to better simulate real-world adversarial attacks.
The results of first priority metrics and the simple and easy-to-implement features prove that this method has higher practical value. 
Thus, ROCKET can be treated as a simple baseline to facilitate future research in this direction.

\paragraph{Secondary Priority Metrics.}
We evaluate secondary priority metrics on Disinformation, Toxic, and Sensitive tasks because successful adversarial samples on other tasks are limited, which will result in inaccurate measures. 
We list the results in Table~\ref{Quality_Evaluation}. 
Our findings are as follows:
\begin{itemize} [topsep=1pt, partopsep=1pt, leftmargin=12pt, itemsep=-2pt]
    \item Considering all attack methods, previously overlooked character-level attacks (e.g., DeepWordBug) achieve great success considering perturbation degree (Levenstein distance) and grammaticality ($\Delta$I). 
    \item While achieving superiority in first priority metrics, ROCKET adds more violent perturbations and breaks the grammaticality more severely. However, as we argue, it's reasonable to trade-off these secondary priority metrics for the first ones. 
    \item Surprisingly, we find that ROCKET crafts more fluent adversarial samples according to the perplexity scores calculated by the language model.
    % which may be counter-intuitive because humans may find the irrelevant appending words meaningless.
    We suspect that the pretraining data that large language models fit on contains so much informal text (e.g.,~Twitter), which may resemble adversarial samples crafted by ROCKET. 
    
\end{itemize}

\subsection{Evaluation on the Defense Side} 
We give the details and results of experiments on the defense side in Appendix~\ref{appendix:experimental_results}.
Table~\ref{defense} shows that DeepWordBug and ROCKET consistently outperform word-level attack methods, indicating that methods on adversarial defense still need to be improved to tackle real-world harmfulness.

\section{Related Work}

\subsection{Adversarial Attack}
% \paragraph{Adversarial Attack. }
% 
Textual adversarial attack methods can be roughly categorized into character-level, word-level, and sentence-level perturbation methods.  

\textbf{Character-level} attacks make small perturbations to the words, including swapping, deleting, and inserting characters \cite{karpukhin-etal-2019-training, gao2018black, ebrahimi-etal-2018-adversarial}. 
These kinds of perturbations are indeed most employed by real-world attackers because of their free of external knowledge and ease of implementation. 
% \yang{add some experimental results and rewrite this paragraph.}
% So, we incorporate some of these modification rules into our simple baseline.  
\textbf{Word-level} attacks can be modeled as a combinatorial optimization problem including finding substitution words and searching adversarial samples. 
Previous work make different practices in these two stages \cite{ren-etal-2019-generating, alzantot-etal-2018-generating, zang-etal-2020-word, li-etal-2020-bert-attack}.
% We don't consider word-level perturbation in our method 
These methods mostly rely on external knowledge bases and are inefficient, rendering them rarely happen in reality. 
\textbf{Sentence-level} attacks paraphrase original sentences to transform the syntactic pattern \cite{iyyer-etal-2018-adversarial}, the text style \cite{qi-etal-2021-mind}, or the domain \cite{wang-etal-2020-cat}.
These kinds of methods rely on a paraphrasing model. Thus, they are also unlikely to happen in reality.  

There also exists some work that cannot be categorized in each of these categories, including multi-granularity attacks~\cite{wang-etal-2020-t3, chen-etal-2021-multi}, token-level attacks~\cite{yuan2021bridge}, and universal adversarial triggers~\cite{wallace-etal-2019-universal, xu-etal-2022-exploring}.

\subsection{Adversarial Defense}
Textual adversarial defense methods can be roughly categorized into five categories based on their strategies, including training data augmentation \cite{si-etal-2021-better}, adversarial training \cite{ren-etal-2019-generating, zang-etal-2020-word, wang2020adversarial, zhu2019freelb, ivgi2021achieving}, preprocessing module \cite{zhou-etal-2019-learning, mozes-etal-2021-frequency, bao-etal-2021-defending}, robust representation learning \cite{jones-etal-2020-robust, liu2020joint, zhou-etal-2021-defense, wang2021natural, pruthi2019combating,tan-etal-2020-mind}, and certified robustness \cite{jia-etal-2019-certified, wang-etal-2021-certified, ye-etal-2020-safer, huang-etal-2019-achieving}.

\label{defenserelated}

\subsection{Security NLP}
The research on security NLP is not only about adversarial attacks in the inference time, 
but also include several other topics that have broad and significant impact in this filed, including privacy attacks \cite{shokri2017membership, pan2020privacy}, backdoor learning \cite{kurita-etal-2020-weight, chen2021textual, cui2022unified}, data poisoning attacks \cite{wallace-etal-2021-concealed, marulli2021exploring}, outlier detection \cite{hendrycks-etal-2020-pretrained, arora2021types}, and so on.
Our \benchmark~can also be employed by some research on security NLP to better reveal the security issues and highlight the practical significance.
\section{Discussion}
\paragraph{Research on Adversarial Attack.}
Note that we don't discredit previous work in this paper. 
Most previous methods are very useful considering different roles of adversarial samples except the security role. For example, although synonym substitution-based methods may not be actually employed by real-world attackers~\cite{ren-etal-2019-generating,zang-etal-2020-word,li-etal-2020-bert-attack}, the adversarial samples, if crafted properly, are very useful for evaluating models' robustness to out-of-distribution data, explaining models' behaviors, and adversarial training.
% Consider word substitution-based algorithms that construct adversarial samples based on synonym substitutions \cite{ren-etal-2019-generating,zang-etal-2020-word,li-etal-2020-bert-attack}.
% These algorithms may not be employed by real-world attackers because their unrealistic assumption on the accessibility to the victim models' confidence scores and inefficiency, which reduces these methods' value in real-world \advintent~research.

But from the perspective of separating roles of adversarial samples, the research significance of adversarial attack methods that assume only the accessibility to the confidence scores of the victim models may be limited. 
When adversarial samples are employed to reveal the security issues, they can only access the models' decisions.
When adversarial samples are used for other purposes, their roles are to help to improve the models at hand. 
In this case, these methods should be granted to have access to the victim model's parameters (i.e. white-box attack) \footnote{Some methods employ “behavioral testing” (black-box testing) even if permission is granted for model parameters \cite{ribeiro-etal-2020-beyond,goel2021robustness}.}. 

Here we only give our considerations of this problem. 
Future research and discussion should go on to refine the problem definition in this field. 

\label{discuss:reflect}

\paragraph{Research on Adversarial Defense.}
% Adversarial defense methods can be employed to better cope with out-of-distribution data.

Adversarial defense methods have two functions, namely making models more robust to out-of-distribution data and resisting malicious adversarial attacks.
Also, we recommend researchers study these two different functions separately. 
For improving models' out-of-distribution robustness, existing work has made many good attempts \cite{si-etal-2021-better, wang2020adversarial}.
However, the impact of existing work on real-world adversarial concerns may be limited because they mostly consider synonym substitution-based attacks that may be less practical in reality \cite{wang2021natural,zhou-etal-2021-defense}.
Thus, we recommend future research on adversarial defense in the security side to consider attack methods that are actually employed by real-world attackers, like the simple ROCKET proposed in this paper.

\paragraph{Research on Security NLP. }
We also conduct a pilot survey on research on the security community.
We find that there exists a research gap between the NLP and the security communities in security research topics. 
While the NLP community puts more emphasis on the methods' novelty, work in the security community usually revolves around actual security scenarios \cite{liao2016seeking, yuan2018reading, wang2020into}. 
Both directions are significant and impactful but a more accurate claim is needed.
We recommend future research on adversarial NLP state clearly what actual goal they aim to achieve (e.g., reveal security concerns or evaluate models' robustness) and develop methods under a reasonable problem definition.

% A keynote delivered in this paper is to recommend the research works on security NLP to situate in reality and consider methods' practical significance.
% We make some attempts and future works are needed to further transform the interesting ideas in academic papers into practical tools in industries.

\section{Conclusion}
In this paper, we rethink the research paradigm in SoadNLP.
% adversarial NLP considering the security role of adversarial samples to reveal practical concerns of NLP models deployed in \advintent~situations.
We identify two major deficiencies in previous work and propose our refinements. 
Specifically, we propose an \advintent~datasets collection \benchmark. 
We then reconsider the actual adversarial goals and reformalize the task.
% focusing on making adjustments to the importance of different evaluation metrics, corresponding to the adversarial goals. 
Next, we propose a simple method summarized from different sources that fulfills real-world attackers' goals. 
We conduct comprehensive experiments on \benchmark~on both the attack and the defense sides. 
Experimental results show the superiority of our method considering the first priority adversarial goals.
The overall experimental results indicate that the current research paradigm in SoadNLP may need to be adjusted to better cope with real-world adversarial challenges.

In the future, we will reconsider and discuss other roles of textual adversarial samples to make this whole story complete.

\subsection*{Ethical Consideration}
In this section, we discuss the potential wider implications and ethical considerations of this paper.

\paragraph{Intended Use.}  In this paper, we construct a \advintent~benchmark, and propose a simple method that can effectively attack real-world SOTA models.
Our motivation is to better simulate real-world adversarial attacks and reveal the practical concerns. This simple method can serve as a simple baseline to facilitate future research on both the attack and the defense sides.
Future work can start from our benchmark and propose methods to address real-world security issues.

% Our motivation is that, as the number of adversarial attack algorithms increased, it became a very important threat in security-related scenarios, so it was imperative to build a security-related Benchmark. In addition, the usual attack methods are based on score or gradient-based methods, but this is not suitable for real scenarios because only decision is available in real scenarios. Our method works well than other method with these benchmarks.
\paragraph{Broad Impact.} 
We rethink the research paradigm in adversarial NLP from the perspective of separating different roles of adversarial samples.
Specifically, in this paper, we focus on the security role of adversarial samples and identify two major deficiencies in previous work. 
For each deficiency, we make some refinements to previous practices. 
In general, our work makes the problem definition in this direction more standardized and better simulate real-world attack situations.  
% Our work gives us an inspiration and a new direction, which is to focus on security-related attacks and defenses in real scenarios. Although the existing attack algorithms are multitudinous, they can not achieve very good results in this scene, but our simple attack algorithm can achieve better results. As an important NLP scenario, security scenario should also be proposed and promoted.
\paragraph{Energy Saving. } We describe our experimental details in Appendix~\ref{appendix:experimental_details} to prevent people from making unnecessary hyper-parameter adjustments and to help researchers quickly reproduce our results.

\section*{Limitation}
\looseness=-1
In experiments, we employ BERT-base as the testbed and evaluate existing textual adversarial attack methods and our proposed ROCKET in our constructed benchmark datasets. 
We only consider one victim model in our experiments because our benchmark includes up to ten datasets and our computing resources are limited. 
Thus, more comprehensive experiments spanning different model architectures and training paradigms are left for future work.

\section*{Acknowledgements}
This work is supported by the National Key R\&D Program of China (No. 2020AAA0106502), Institute Guo Qiang at Tsinghua University  and NExT++ project from the National Research Foundation, Prime Minister’s Office, Singapore under its IRC@Singapore Funding Initiative.

Yangyi Chen made the original research proposal and wrote the paper. 
Hongcheng Gao conducted experiments and helped to organize the paper.
Ganqu Cui and Fanchao Qi revised the paper and participated in the discussion. Longtao Huang, Zhiyuan Liu and Maosong Sun advised the project.

\bibliography{anthology,custom}
\bibliographystyle{acl_natbib}

\appendix

\label{sec:appendix}
\section{Survey on Previous Work}
We conduct a survey on previous adversarial attack methods about the specific tasks and datasets they employ in their evaluation. The results are listed in Table~\ref{previousworks}.

\section{Task Description}
\label{appendix:task}
The task statistics are listed in Table~\ref{dataset_description}. We give the task descriptions and our motivations to choose these tasks below.

\subsection{Misinformation}
Words in news media and political discourse have considerable power in shaping people’s beliefs and opinions. 
As a result, their truthfulness is often compromised to maximize the impact on society~\cite{zhang2020overview,zhou2019fake,fonseca2016measuring}.  
We generally believe that fake news is caused by objective factors such as misdeclarations, misdescriptions, or misuse of terminology.
% Therefore, we deem the detection of such fake news caused by objective factors as the misinformation detection task. 
And this task is to detect misinformation that contains deceived or unverified information,
including rumors, misreported, and satirical news. 

\subsection{Disinformation}
In addition to misinformation caused by objective reasons, there is also a type of fake information caused by subjectively distorting facts. 
This type of information mainly concentrates on online comments and reviews in online shopping malls and online restaurant/hotel reservation websites to lure customers into consumption~\cite{mukherjee2013fake,sun2016exploiting,patel2018survey}.
We define such a task as disinformation detection.
In general, this task is dedicated to identifying deliberate fabrication of facts, including (1) Artificial comments reversing the black and white; 
(2) Generated nonexistent information.

\subsection{Toxic}
The rapid growth of information in social networks such as Facebook, Twitter, and blogs makes it challenging to monitor what is being published and spread on social media.  
Abusive comments are widespread on social networks, including cyberbullying, cyberterrorism, sexism, racism, and hate-speech. 
Thus, the primary objective of toxic detection is to identify toxic contents in the web, which is an essential ingredient for anti-bullying policies and protection of individual rights on social media~\cite{pereira2019detecting,bosco2018overview,watanabe2018hate,risch2020toxic}.

\subsection{Spam}
In recent years, unwanted commercial bulk emails have become a huge problem on the internet.  
Spam emails prevent the user from making good use of time.
More importantly, some spam emails contain fraud and phishing messages that can also cause financial damage to users~\cite{fonseca2016measuring}. 
The Spam Classification Tasks is to detect spam information including scams, harassment, advertising, and promotion in Emails, SMS, and even chat messages to avoid unnecessary losses for users~\cite{cormack2008email}.

\begin{table}[]
\centering
\resizebox{.45\textwidth}{!}{
\begin{tabular}{lll}
\toprule[1.2pt]
\multicolumn{1}{l}{\textbf{Work}} & \multicolumn{1}{l}{\textbf{Task}}                                                                       & \multicolumn{1}{l}{\textbf{Dataset} }                                                                                 \\ \hline
\cite{alzantot-etal-2018-generating}                        & {\color[HTML]{000000} SA; NLI.}                                                                 & {\color[HTML]{000000} IMDB; SNLI.}                                                                           \\ \hline
\citep{ren-etal-2019-generating}                      & {\color[HTML]{000000} \begin{tabular}[c]{@{}l@{}}SA; NC; \\ Topic classification.\end{tabular}} & {\color[HTML]{000000} \begin{tabular}[c]{@{}l@{}}IMDB; AG; \\ Yahoo! Answers.\end{tabular}}                  \\ \hline
\cite{jin2020bert}                & {\color[HTML]{000000} \begin{tabular}[c]{@{}l@{}}SA; NLI; NC;\\ \color[HTML]{FE0000}Fake News.\end{tabular}}        & {\color[HTML]{000000} \begin{tabular}[c]{@{}l@{}}IMDB; Yelp; MR; SNLI; \\ MNLI; AG;  \color[HTML]{FE0000}Fake News.\end{tabular}} \\ \hline
\cite{zang-etal-2020-word}                       & {\color[HTML]{000000} SA; NLI.}                                                                 & {\color[HTML]{000000} IMDB; SST-2; SNLI.}                                                                    \\ \hline
\cite{li-etal-2020-bert-attack}               & {\color[HTML]{000000} \begin{tabular}[c]{@{}l@{}}SA; NLI; NC; \\\color[HTML]{FE0000} Fake News.\end{tabular}}       & {\color[HTML]{000000} \begin{tabular}[c]{@{}l@{}}IMDB;  Yelp; SNLI; \\ MNLI; AG; \color[HTML]{FE0000} Fake News.\end{tabular}}    \\ \hline
\cite{maheshwary2021generating}                    & {\color[HTML]{000000} SA; NLI; NC.}                                                             & {\color[HTML]{000000} \begin{tabular}[c]{@{}l@{}}IMDB; Yelp; MR; SNLI; \\ MNLI; AG; Yahoo.\end{tabular}}     \\ \hline
\cite{iyyer-etal-2018-adversarial}                      & {\color[HTML]{000000} SA; NLI.}                                                                 & {\color[HTML]{000000} SST-2; SICK.}                                                                          \\ \hline
\cite{chen-etal-2021-multi}                      & {\color[HTML]{000000} SA; NLI;  NC,}                                                            & {\color[HTML]{000000} SST-2; MNLI; AG.}                                                                      \\ \hline
\cite{qi-etal-2021-mind}                  & {\color[HTML]{000000} \begin{tabular}[c]{@{}l@{}}SA; NC;  \\ \color[HTML]{FE0000} Hate-Speech.\end{tabular}}         & {\color[HTML]{000000} \begin{tabular}[c]{@{}l@{}}SST-2; AG; \\ \color[HTML]{FE0000} Hate-Speech.\end{tabular}}                    \\ \hline
\cite{li-etal-2021-contextualized}                     & {\color[HTML]{000000} SA; NLI; NC.}                                                             & Yelp; MNLI;  QNLI; AG.                                                                                       \\ \hline
\cite{li-etal-2021-contextualized}                     & {\color[HTML]{000000} SA; NLI; NC.}                                                             & Yelp; MNLI;  QNLI; AG.                                                                                       \\ \hline

\cite{yuan2021bridge}                & {\color[HTML]{000000} SA; NLI;  NC,}                                                                      & {\color[HTML]{000000} SST-2; MNLI; AG.}                                                                                                  \\ \bottomrule[1.2pt]
\end{tabular}
}
\caption{\label{previousworks} Survey on previous work. \textbf{SA} stands for sentiment analysis. \textbf{NC} stands for news classification. Adversarial-oriented tasks and datasets are highlighted in \color[HTML]{FE0000} \textbf{red}\color[HTML]{000000}.}

\end{table}

% \usepackage{multirow}
\begin{table*}[]
\centering
\resizebox{.96\textwidth}{!}{\begin{tabular}{cccccccccc}
\toprule[1.2pt]
\multirow{2}{*}{\textbf{Task}}  & \multirow{2}{*}{\textbf{Dataset}} & \multicolumn{4}{c}{\underline{\textbf{Unbalanced}}} &                      & \multicolumn{3}{c}{\underline{\textbf{Balanced}}} \\
                                &                                   & |Train|  & |Test| & Ave. Length & Ratio &                      & |Train|    & |Test|   & Ave. Length   \\ \hline
\multirow{2}{*}{Misinformation}           & LUN                               & 36148    & 15492  & 535.33      & 0.79  &                      & 14906      & 6454     & 499.49        \\
                                & SATNews                           & 145677   & 37221  & 702.51      & 0.09  &                      & 25264      & 7202     & 646.65        \\ \hline
\multirow{2}{*}{Disinformation} & Amazon-LB                         & 17902    & 8609   & 99.01       & 0.49  &                      & 17434      & 8522     & 100.13        \\
                                & CGFake                            & 28290    & 12130  & 67.48       & 0.50  &                      & 28290      & 12130    & 67.48         \\ \hline
\multirow{2}{*}{Toxic}          & HOSL                              & 17348    & 7435   & 14.12       & 0.83  &                      & 5832       & 2494     & 14.32         \\
                                & Jigsaw2018                            & 159560   & 63978  & 66.52       & 0.10  &                      & 30587      & 12180    & 58.42         \\ \hline
\multirow{2}{*}{Spam}           & Enron                             & 17774    & 7918   & 311.47      & 0.46  &                      & 16159      & 7277     & 311.53        \\
                                & SpamAssassin                       & 3766     & 3774   & 291.75      & 0.28  &                      & 2081       & 2066     & 308.50        \\ \hline
\multirow{2}{*}{Sensitive Information}      & EDENCE                            & 105376   & 22577  & 22.45       & 0.24  &                      & 51098      & 10328    & 21.79         \\
                                & FAS                               & 60470    & 16272  & 27.65       & 0.31  & \multicolumn{1}{l}{} & 33814      & 13294    & 29.27         \\ \bottomrule[1.2pt]
\end{tabular}}
\caption{\label{dataset_description} Datasets statistics. The ratio refers to the proportion of fake/hate/spam/sensitive samples in corresponding datasets. }

\end{table*}

\subsection{Sensitive Information}
Text documents shared across third parties or published publicly contain sensitive information by nature. 
Detecting sensitive information in unstructured data is crucial for preventing data leakage.
This task is to detect sensitive information including intellectual property and product progress from companies, trading and strategic information of public institutions and organizations, and private information of individuals~\cite{berardi2015semi,chow2008detecting,grechanik2014redacting}.

\section{Definition of Validity}
\label{appendix:validity_def}

In general, the validity metric is to measure the preservation of adversarial meaning in the crafted adversarial samples. 
The adversarial meaning is task-specific and should be considered differently. 
So, the validity definition is relevant to the specific adversarial goal in the specific \advintent~task. 
In our \benchmark, the adversarial meanings are exaggerated and satirical contents (Misinformation), inauthentic and untrue comments (Disinformation), abusive language (Toxic), illegal or time-wasting messages (Spam), and sensitive information embedded in common comments (Sensitive Information). 
So, the ultimate goal of attackers is to spread the adversarial meaning, no matter how many perturbations attackers introduce to other unrelated content.

% We discuss the validity definition of tasks considered in our \benchmark: 

% \begin{itemize} [topsep=1pt, partopsep=1pt, leftmargin=12pt, itemsep=-2pt]
% 	\item \textbf{Misinformation: } The ultimate goal of attackers is to spread the exaggerated and satirical information to the public. So, valid adversarial samples should retain the exaggerated and satirical contents, no matter how much perturbations attackers introduce to other unrelated content. 
% 	\item \textbf{Disinformation:} The validity definition of disinformation detection is almost the same as Misinformation detection. 
% 	\item \textbf{Toxic:} The ultimate goal of attackers is to spread the abusive message to the social media. 
% 	So, valid adversarial samples should retain the abusive meaning contained in the original sample.
	
% 	\item \textbf{Spam:} The ultimate goal of attackers is to spread some illegal or time-wasting messages to the public through emails or chat platforms. 
% 	\item \textbf{Sensitive Information:}
% \end{itemize}

\section{Real-world Adversarial Attack}
\label{appendix:real_word_case}

\begin{figure*}[ht]
\centering
\includegraphics[width=0.9\textwidth]{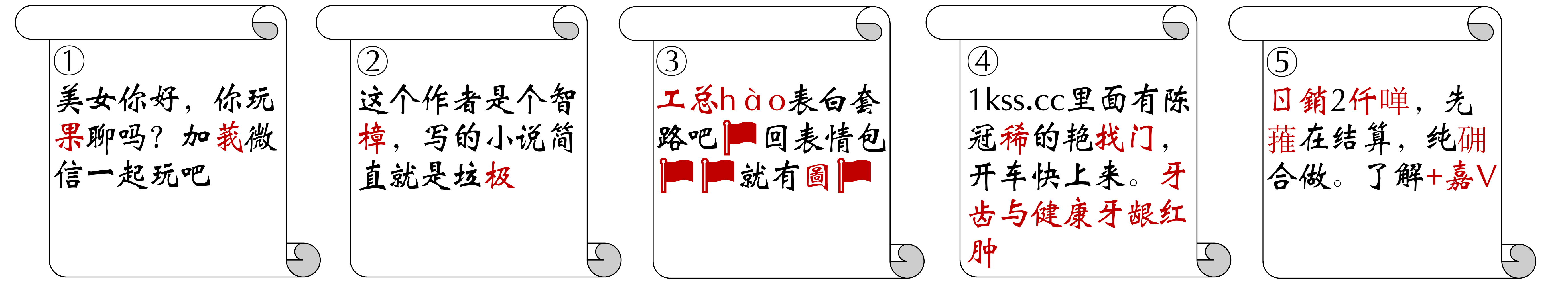}
\caption{Real-world cases of adversarial attacks. Adversarially modified content is highlighted in \color[HTML]{FE0000} \textbf{red}\color[HTML]{000000}.}
\label{fig:real_world}
\end{figure*}

We give some real-world adversarial cases collected from social media in Figure~\ref{fig:real_world}. 
Although these cases are written in Chinese, the perturbation rules are general and widely applicable. 
We can see that case-1, case-2, and case-5 also employ character-level perturbations, including substitution, deletion, and insertion.
Besides, case-3 and case-4 employ the strategy of adding irrelevant and distracting words to the original sample. 
These samples can be easily comprehended by humans but easily fool the detection system. 
We employ these strategies in our simple method to simulate real-world adversarial attacks.

\begin{figure*}[ht]
\centering
\includegraphics[width=0.95\textwidth]{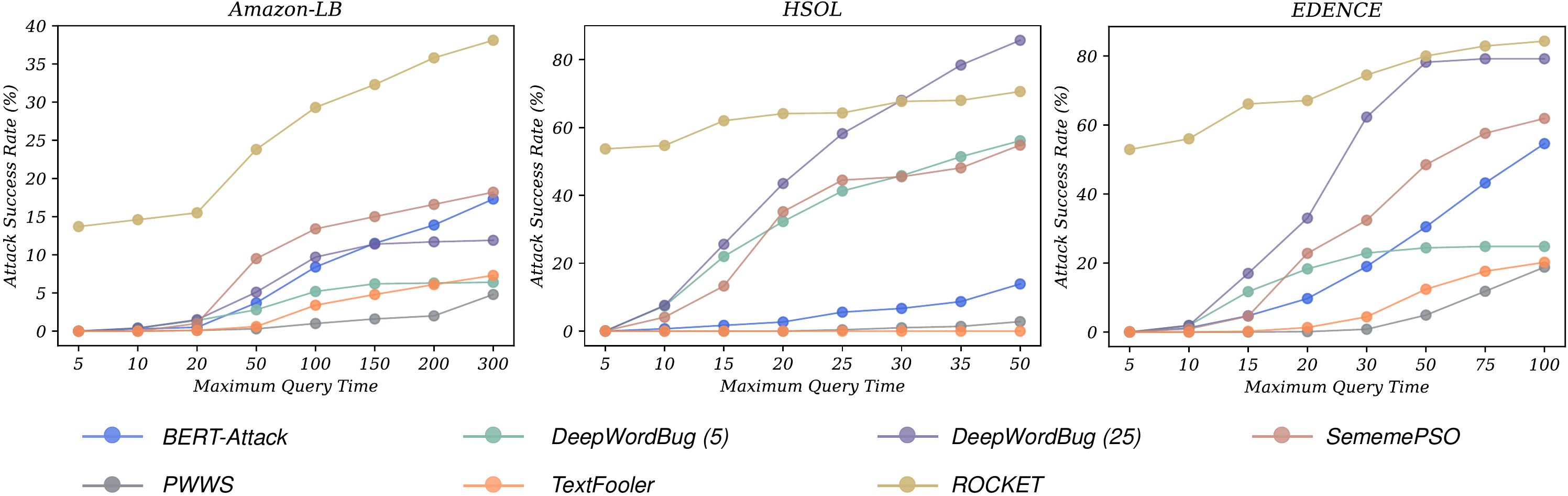}
\caption{Attack success rate under the restriction of maximum query times.}
\label{fig:querytime}
\end{figure*}

\section{Experimental Results}
\label{appendix:experimental_results} 
\subsection{Attack Efficiency}
Figure~\ref{fig:querytime} shows the results of the attack success rate under the restriction of maximum query times.

\subsection{Secondary Priority Metrics}
We list the results of secondary priority metrics in Table~\ref{Quality_Evaluation}.
\begin{table*}[]
\centering
\resizebox{.95\textwidth}{!}{

\begin{tabular}{lccclccclccc}
\toprule[1.2pt]
\textbf{Task}                            & \multicolumn{3}{c}{\textbf{Disinformation}}    & \multicolumn{1}{c}{\textbf{}} & \multicolumn{3}{c}{\textbf{Toxic}}               & \multicolumn{1}{c}{\textbf{}} & \multicolumn{3}{c}{\textbf{Sensitive Information}}          \\ \hline
\multirow{2}{*}{\textbf{Method~|~Dataset}} & \multicolumn{3}{c}{\underline{Amazon-LB}}                  &                               & \multicolumn{3}{c}{\underline{HSOL}}                         &                               & \multicolumn{3}{c}{\underline{EDENCE} }                     \\
                                         & Levenstein     & $\Delta$I            & \%PPL         & \textbf{}                     & Levenstein     & $\Delta$I             & \%PPL          & \textbf{}                     & Levenstein    & $\Delta$I             & \%PPL          \\ \hline
TextFooler                               & 25.68          & 0.57          & \textbf{1.79} &                               & 10.83          & 0.03           & 0.89           &                               & 11.86         & 0.10           & 2.51           \\
PWWS                                     & 95.89          & 1.27          & 3.69          &                               & 18.13          & 0.06           & 2.49           &                               & 30.06         & 0.21           & 5.84           \\
BERT-Attack                              & 117.31         & 0.41          & 6.60          &                               & 22.13          & 0.06           & 3.13           &                               & 24.78         & 0.07           & 3.69           \\
SememePSO(maxiter=100)                   & 28.58          & 0.64          & 1.93          &                               & 13.42          & 0.06           & 2.17           &                               & 14.71         & 0.09           & 3.35           \\
DeepWordBug(power=5)                     & \textbf{18.86} & 0.33          & 4.19          &                               & \textbf{9.14}  & -0.01          & 2.57           &                               & \textbf{6.66} & -0.02          & 5.13           \\
DeepWordBug(power=25)                    & 34.43          & \textbf{0.23} & 4.27          &                               & 18.96          & \textbf{-0.09} & 1.32           &                               & 19.57         & \textbf{-0.04} & 2.41           \\
\textbf{ROCKET}                          & 82.93          & 1.01          & 1.81          &                               & 1083.42        & 44.99          & \textbf{-0.98} &                               & 85.75         & 4.99           & \textbf{-0.46} \\ \hline
\multicolumn{12}{l}{}                                                                                                                                                                                                                                          \\ \hline
\multirow{2}{*}{\textbf{Method~|~Dataset}} & \multicolumn{3}{c}{\underline{CGFake}}                     &                               & \multicolumn{3}{c}{\underline{Jigsaw2018} }                      &                               & \multicolumn{3}{c}{\underline{FAS}}                         \\
                                         & Levenstein     & $\Delta$I            & \%PPL         & \textbf{}                     & Levenstein     & $\Delta$I             & \%PPL          & \textbf{}                     & Levenstein    & $\Delta$I             & \%PPL          \\ \hline
TextFooler                               & 16.62          & 0.23          & 2.12          &                               & 14.64          & 0.09           & 1.21           &                               & 13.12         & 0.05           & 2.45           \\
PWWS                                     & 101.47         & 1.11          & 4.69          &                               & 52.76          & 0.38           & 4.19           &                               & 48.93         & 0.23           & 5.04           \\
BERT-Attack                              & 82.85          & 0.48          & 11.41         &                               & 30.65          & 0.05           & 3.33           &                               & 53.20         & 0.10           & 6.12           \\
SememePSO(maxiter=100)                   & 23.77          & 0.31          & 3.15          &                               & 18.23          & 0.11           & 3.42           &                               & 15.55         & 0.03           & 2.37           \\
DeepWordBug(power=5)                     & \textbf{10.01} & 0.05          & 5.16          &                               & \textbf{10.79} & 0.03           & 3.45           &                               & \textbf{6.65} & -0.02          & 3.64           \\
DeepWordBug(power=25)                    & 29.49          & \textbf{0.00}    & 4.13          &                               & 20.49          & \textbf{-0.05} & 2.27           &                               & 18.33         & \textbf{-0.03} & 2.44           \\
\textbf{ROCKET}                          & 38.27          & 1.03          & \textbf{1.66} &                               & 1084.44        & 44.99          & \textbf{-0.96} &                               & 97.95         & 5.01           & \textbf{0.34}  \\ \bottomrule[1.2pt]
\end{tabular}

}

\caption{\label{Quality_Evaluation} Results of secondary priority metrics considering perturbation degree and grammaticality. }

\end{table*}

\subsection{Evaluation on the Defense Side}

\begin{table*}[]
\centering
\resizebox{.78\textwidth}{!}{\begin{tabular}{lcccccc}
\toprule[1.2pt]
\textbf{Task}                            & \multicolumn{2}{c}{\textbf{Disinformation}} & \multicolumn{2}{c}{\textbf{Toxic}}          & \multicolumn{2}{c}{\textbf{Sensitive Information}} \\ \hline
\multirow{2}{*}{\textbf{Method~|~Dataset}} & \multicolumn{2}{c}{\underline{Amazon-LB}}               & \multicolumn{2}{c}{\underline{HSOL}}                    & \multicolumn{2}{c}{\underline{EDENCE}}                         \\
                                         & $R_{res}$(\%)    & $R_{det}$(\%)   & $R_{res}$(\%)    & $R_{det}$(\%)   & $R_{res}$(\%)        & $R_{det}$(\%)      \\ \hline
TextFooler                               & 57.89                & 73.68                & 85.00                & 86.00                & 43.40                    & 53.19                   \\
PWWS                                     & 49.28                & 69.57                & 87.76                & 89.80                & 46.62                    & 58.82                   \\
BERT-Attack                              & 18.29                & 43.90                & 17.02                & 33.51                & 25.31                    & 33.93                   \\
SememePSO(maxiter=100)                   & 52.54                & 79.66                & 79.70                & 83.16                & 60.28                    & 66.92                   \\
DeepWordBug(power=5)                     & \textbf{0.00}        & \textbf{10.00}       & 6.05                 & 12.46                & 4.41                     & 7.05                    \\
DeepWordBug(power=25)                    & 1.49                 & 11.94                & \textbf{0.94}        & \textbf{1.41}        & \textbf{0.88}            & \textbf{1.88}           \\
\textbf{ROCKET}                          & 14.15                & 35.05                & 12.75                & 30.03                & 8.11                     & 15.75                   \\ \hline
                                         & \multicolumn{1}{l}{} & \multicolumn{1}{l}{} & \multicolumn{1}{l}{} & \multicolumn{1}{l}{} & \multicolumn{1}{l}{}     & \multicolumn{1}{l}{}    \\ \hline
\multirow{2}{*}{\textbf{Method~|~Dataset}} & \multicolumn{2}{c}{\underline{CGFake}}                  & \multicolumn{2}{c}{\underline{Jigsaw2018}}              & \multicolumn{2}{c}{\underline{FAS}}                            \\
                                         & $R_{res}$(\%)    & $R_{det}$(\%)   & $R_{res}$(\%)    & $R_{det}$(\%)   & $R_{res}$(\%)        & $R_{det}$(\%)      \\ \hline
TextFooler                               & 4.42                 & 7.73                 & 22.58                & 33.06                & 56.90                    & 63.79                   \\
PWWS                                     & 5.10                 & 7.43                 & 68.37                & 70.92                & 61.92                    & 71.23                   \\
BERT-Attack                              & \textbf{0.11}        & 6.37                 & 15.26                & 22.37                & 27.89                    & 37.91                   \\
SememePSO(maxiter=100)                   & 9.43                 & 13.73                & 50.77                & 61.20                & 71.29                    & 75.37                   \\
DeepWordBug(power=5)                     & 1.69                 & 13.08                & 1.67                 & 12.53                & 1.23                     & 7.35                    \\
DeepWordBug(power=25)                    & 1.17                 & 8.45                 & \textbf{0.35}        & \textbf{3.47}        & \textbf{0.13}            & \textbf{1.80}           \\
\textbf{ROCKET}                          & 0.31                 & \textbf{2.49}        & 7.00                 & 22.26                & 5.64                     & 13.11                   \\ \bottomrule[1.2pt]
\end{tabular}
}

\caption{\label{defense} Results on the defense side.}

\end{table*}
The results are shown in Table~\ref{defense}.
We employ the SOTA defense method proposed in the NLP community~\cite{mozes-etal-2021-frequency}.
% $briefly describe the defense method, how it functions$
This method identifies word substitutions by the frequency difference between the substituted word and its corresponding substituted word.  
The frequency distribution of words is obtained on the training set, and the detector is tuned on the validation set. 
Then, the detector can be employed to identify and restore adversarial samples in the inference time.

For each attack method, we input $N$ adversarial samples (successfully attack the model) to the trained detector to obtain the number of samples detected as adversarial samples ($n_{det}$) and the number of samples successfully restored ($n_{res}$). 
Then the \textbf{detection rate} ($R_{det}$) and \textbf{restored rate} ($R_{res}$) are calculated according to the formula:
\begin{equation}
\label{eq}
\left\{
\begin{aligned}
R_{res}=\frac{n_{res}}{N} \\
R_{det}=\frac{n_{det}}{N}
\end{aligned}
\right.
\end{equation}

% We employ the SOTA defense method proposed in the NLP community~\cite{mozes-etal-2021-frequency}.We employ the SOTA defense method proposed in the NLP community~\cite{mozes-etal-2021-frequency}.
% We employ the SOTA defense method proposed in the NLP community~\cite{mozes-etal-2021-frequency}.

\section{Experimental Details}
\label{appendix:experimental_details}

\begin{table*}[]
\centering
\resizebox{.95\textwidth}{!}{
\begin{tabular}{lccccc}
\toprule[1.2pt]
\textbf{Parameter | Task} & \multicolumn{1}{l}{\textbf{Misinformation}} & \multicolumn{1}{l}{\textbf{Disinformation}} & \multicolumn{1}{l}{\textbf{Toxic}} & \multicolumn{1}{l}{\textbf{Spam}} & \multicolumn{1}{l}{\textbf{Sensitive Information}} \\ \hline
Distracting Word        & reuters                                     & up                                          & peace                              & \textgreater{}                    & any                                                \\
Prefix Number           & 5                                           & 3                                           & 0                                  & 10                                & 0                                                  \\
Postfix Number          & 30                                          & 8                                           & 180                                & 30                                & 20                                                 \\
Batch Size               & 8                                           & 6                                           & 4                                  & 6                                 & 3                                                  \\
Epoch                   & 3                                           & 3                                           & 3                                  & 2                                 & 3                                                  \\
Max Modification Number   & 100                                         & 100                                         & 180                                & 30                                & 100                                                                                \\ \bottomrule[1.2pt]
\end{tabular}
}

\caption{\label{para} Hyper-parameters of ROCKET on each task.}

\end{table*}

For the sake of calculation speed and fairness, we truncate all sentences to the first 480 words. 
Then, we empirically set the hyper-parameters including distracting words, the insertion number of distracting words at the beginning and end of sentences, perturbation batch size, and perturbation epochs according to the attack performance and preservation of adversarial meaning. 
We only attack the original content in sentences, leaving out adversarial content introduced by our perturbations. 
The comprehensive settings of hyper-parameters are shown in Table~\ref{para}.

Here we give our intuition for choosing distracting words for each task.
For misinformation detection, we find that newspaper names often appear at the beginning or end of the news. 
So, we insert a few "Reuters" before and after the sentence without affecting the validity of the main content.
For Disinformation detection, adding several encouraging words "up" does not affect the judgment of the authenticity of the comments, so we use a number of "up" as parenthetical words. 
For toxic detection, we need to use friendly and harmonious words to fool detectors. 
So, we insert many "peace" to the sentences. 
For spam detection, we find that ">" sometimes appears in emails to separate the text.
So, we use a large number of them as inserted words, which doesn't affect the nature of the original sentence. 
For sensitive information detection, we employ "any" as we find that samples that are often classified as non-sensitive contain adverbs at the end.

% Considering that attack on a sentence can attack only the body sentence that does not have a new word inserted, or the sentence after the new word is inserted, we set up the "only attack body" to indicate whether to attack only the body part uninserted new word. However, the experimental results show that the best parameter is "TRUE", which means only the trunk part is attacked. Increasing batchesize can reduce the number of queues, increasing epoch can increase ASR, and we get the recommended parameters in the Table~\ref{para} by parameters tuning.

% We train the model for adversarial attack, recording the attack success rate and query times. 
% We use BERT-uncased as the victim model, considering that the attack algorithms in OpenAttack is case-insensitive. 

% Finally we use extra datasets to make hyper-parameter adjustments to each type of task in our method, including inserting words, number of inserts before sentences, number of end-of-sentence insertions, batchsize, and epoch. 

\section{Human Evaluation Details}
\label{appendix:human_eval_details}
We set up a human evaluation to further evaluate the validity of adversarial samples. 
We choose the disinformation and toxic detection tasks because the validity definitions are clear and can be easily understood by annotators.
For each task, we consider 2 corresponding datasets and sample 100 original and adversarial samples pairs for each attack method.
For each pair, we ask 3 human annotators to evaluate whether the adversarial meaning is preserved in the adversarially crafted sample (validity). 
They need to give a validity score from 0-2 for each pair, where 2 means that the adversarial meaning has been perfectly preserved, 1 means that the sentence meaning is ambiguous but may still preserve some adversarial meaning, and 0 means that the crafted adversarial sample don't preserve any adversarial meaning in the original sample. 
We use the voting strategy to produce the annotation results of validity for each adversarial sample.
Then we average the scores for all 100 samples in each task as the final validity score for each attack method. 
The results are shown in Table~\ref{human_validity}.

\end{document}